\documentclass[11pt]{article}
\usepackage{geometry}
\usepackage{coling2020}
\usepackage{times}
\usepackage{url}
\usepackage{latexsym}
\usepackage{microtype}
\usepackage{graphicx}
\usepackage{float}
\usepackage{multirow}

\colingfinalcopy 

\title{NUAA-QMUL at SemEval-2020 Task 8:  Utilizing BERT and DenseNet for Internet Meme Emotion Analysis}

\author{Xiaoyu Guo\\
  Nanjing University of\\
  Aeronautics and Astronautics\\
  {\tt xy\_deepa@163.com} \\\And
  Jing Ma \\
  Nanjing University of\\
  Aeronautics and Astronautics\\
  {\tt majing5525@126.com} \\\And 
  Arkaitz Zubiaga\\
  Queen Mary\\
  University of London\\  
  {\tt a.zubiaga@qmul.ac.uk} \\}

\date{}

\begin{document}
\maketitle
\begin{abstract}
  This paper describes our contribution to SemEval 2020 Task 8: Memotion Analysis. Our system learns multi-modal embeddings from text and images in order to classify Internet memes by sentiment. Our model learns text embeddings using BERT and extracts features from images with DenseNet, subsequently combining both features through concatenation. We also compare our results with those produced by DenseNet, ResNet, BERT, and BERT-ResNet. Our results show that image classification models have the potential to help classifying memes, with DenseNet outperforming ResNet. Adding text features is however not always helpful for Memotion Analysis.
\end{abstract}

\section{Introduction}
\label{intro}

%
%
\blfootnote{
    
    

    
    \hspace{-0.3cm}  
    This work is licensed under a Creative Commons 
    Attribution 4.0 International Licence.
    Licence details:
    \url{http://creativecommons.org/licenses/by/4.0/}.\\
    Source code for our model and for the training procedure is published on \url{https://github.com/xxxxxxxxy/Memotion-Analysis}.
    
    
}

The growing ubiquity of Internet memes on social media platforms such as Facebook, Instagram, and Twitter reinforces the importance of processing multimodal content. Despite memes being so deeply ingrained in the Internet culture, it is still hard to leverage AI methods to understand the underlying meanings expressed by memes. This, in turn, makes it challenging to identify certain kinds of content in these memes, such as malicious information posted on social media. To tackle this problem, SemEval 2020 Task 8 held a meme emotion analysis task\cite{chhavi2020memotion}. It consisted of three sub-tasks: (1) predicting the sentiment of Internet memes as positive, neutral or negative, (2) identifying the type of humor expressed as sarcastic, humorous, offensive or motivation, and (3) quantifying the extent to which a particular emotion is being expressed.

Internet memes can comprise a variety of modalities such as textual, visual, and audio. Adhering to the task guidelines at hand, we approach the problem by considering only the image-with-caption class of meme, which is commonly found on social media. This paper describes a system that combines BERT (Bidirectional Encoder Representation from Transformers) and DenseNet (Dense Convolutional Network) to classify Internet memes. We find that the performance of Multi-modal Deep Neural Network is better than one model Deep Neural Network and the performance of ResNet (Residual Network) is worse than DenseNet.

This paper is organized as follows. In section Background, we describe Memotion Analysis task, the characteristics of the Internet memes, prior work on multi-modal classification, prior research exploring data analysis approaches to study memes, and our contributions in this paper. Then in Section 3, we propose BERT-DenseNet, a deep learning model to classify memes. We then use the models to study memes in task A, task B, and task C in Section 4. In Section 5, we display the results of our experiments, including comparative experiments. Finally, we conclude the findings of this research and suggest directions for future work.
\section{Background}

Task A of this competition consists of classifying an Internet meme as positive, negative, or neutral. Task B consists of identifying whether an Internet meme is sarcastic, humorous, offensive, or motivation. A meme can have more than one category in this case. Task C consists of quantifying the scales of every type of humor. The system should classify it as positive for task A, identify it as sarcastic, humorous, not offensive and not motivational for task B, and quantify it as general (the scale of sarcasm), hilarious (the scale of humor) and not offensive (the scale of offensive) for task C.

A dataset with 6,992 annotated memes is used in this task \cite{chhavi2020memotion}. The dataset contains a set of meme images, the corresponding textual content OCR’d and manually corrected, and the corresponding class. A meme is not only about the funny picture, the Internet culture, or the emotion that passes along, but also about the richness and uniqueness of its language: It is often highly structured with a special writing style and forms interesting and subtle connotations that resonate among the readers \cite{wang2015can}. For example, the Boromir memes with the caption of “one does not simply…” are derived from the movie Lord of the Rings. It shows that the same example can lead to the production of several memes making use of the same background image but different texts, which gives it different meanings and contexts.

Although the language of Internet memes is a relatively new research topic, our work is broadly related to studies on multi-modal classification \cite{Kalva2007,merdivan2019image,Zhou2016,Zhu2019} and multi-modal sentiment analysis \cite{Baecchi2016,hu2018multimodal,Zhongqiang2017}. Most recently, Sun et al. \shortcite{sun2019multi} learned text embeddings using BERT, the current state-of-the-art in text encoders. However, none of the above studies have applied multi-modal approaches for memes sentiment analysis.

Previous research on memes mainly focuses on the automatic generation of memes \cite{costa2015reality,oliveira2016one,peirson2018dank,veale2015twitter}, but not on the analysis as we do in this task. In a typical automatic generation work, Wang \& Wen \shortcite{wang2015can} combined textual and visual information for predicting and generating popular meme descriptions. Beskow, Kumar, \& Carley \shortcite{Beskow2020} detected and characterized Internet memes with multi-modal deep learning. However, fewer studies focus on sentiment and emotion analysis of memes. One of the few works who did focus on sentiment analysis is that by Costa et al. \shortcite{costa2015reality}, who developed a Maximum Entropy classifier for memes sentiment classification, which was however in an early stage of development.

Based on previous work, we take a multi-modal approach to do memes sentiment analysis. We propose to use Keras functional API\footnote{\url{https://keras.io/getting-started/functional-api-guide/}} to model the multimodal dependencies among images, text, and corresponding labels.

We then introduce BERT \cite{devlin2019bert} and DenseNet \cite{huang2017densely} to the model. In empirical experiments, we show that our model outperforms strong discriminative baselines by large margins except for the second sub-task (B). Our contributions are three-fold:

\begin{enumerate}
    \item We perform  sentiment analysis of memes combining NLP, CV, and deep learning techniques, and show that combining the visual and textual signals helps classifying memes by sentiment;
    \item We find that the performance of DenseNet is better than ResNet, and provide an analysis of this result;
    \item Our proposed deep learning model outperforms competitive baselines except for task B.
\end{enumerate}

\section{System overview}

\begin{figure}[htb] 
    \centering 
    \includegraphics[width=1\textwidth]{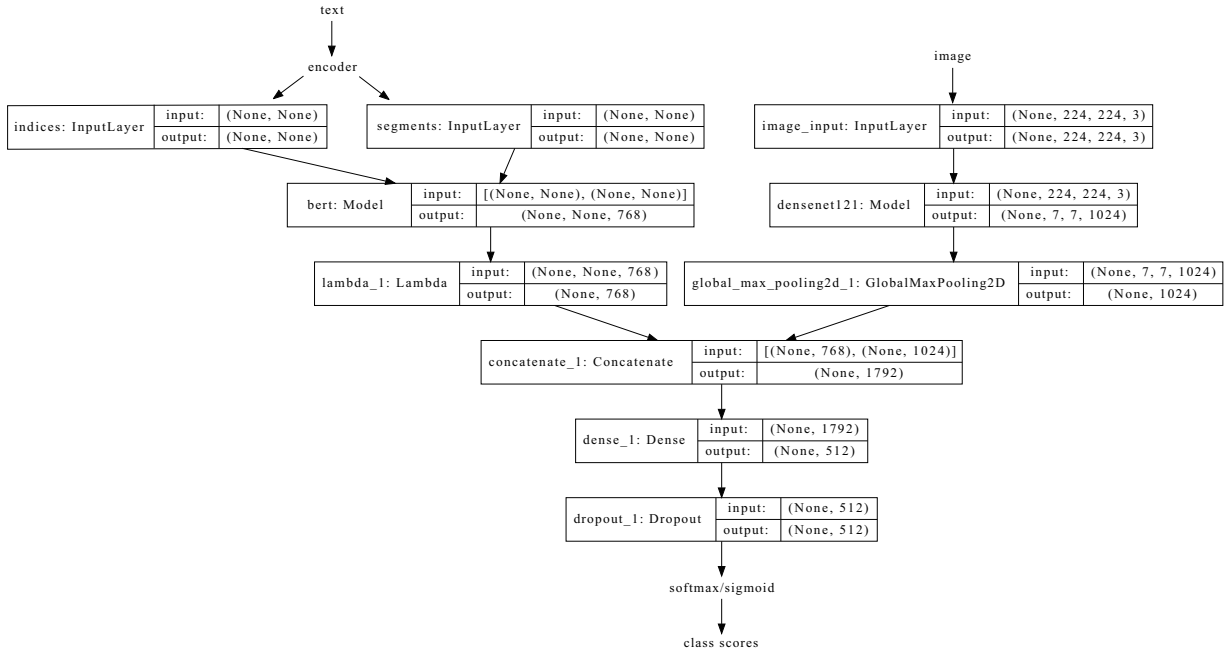} 
    \caption{The BERT-DenseNet model. Softmax activation is used for tasks A and C. Sigmoid activation is for task B.} 
    \label{figure3} 
\end{figure}

We approach the task of Memotion Analysis as eight classification sub-tasks (one for task A, four for task B and three for task C). Four classification sub-tasks from task B are binary but others are multiclass. Our model is depicted in Figure \ref{figure3}, with inputs at the top and outputs at the bottom. In a nutshell, our model combines BERT that processes the caption text and DenseNet that extracts features from images.

\textbf{Data Pre-processing}: Firstly, we delete quotation marks because this would raise an error when reading CSV files. Secondly, we copy text from OCR\_extracted\_text to corrected\_text when the value of corrected\_text is null and the value of OCR\_extracted\_text is not null. We check the corresponding image and add the text to corrected\_text if the value of OCR\_extracted\_text and corrected\_text are both null. Thirdly, we use WordPiece tokenisation, which can break a word into more than one sub-word. This kind of tokenisation is beneficial for memes because they contain several out of vocabulary words.

\textbf{Text Encoding}: To encode text in the data, we use pre-trained BERT \cite{devlin2019bert}, which is a transformer-based language model that conditions jointly on the left and right of a given word. Typically, the BERT encoder is fine-tuned for a particular task by learning an additional task-specific weight layer. In this paper, the layers of BERT, pre-trained on a large corpus of Wikipedia+Book corpus data, are set to be trainable, and additional fine-tuning is not performed. The reason why we choose this encoder is motivated by the success of BERT in achieving state-of-the-art results in several NLP tasks such as sentiment analysis, machine reading comprehension, question-answering, etc. The input text in Section 4 is encoded using BERT-cased\_L-12\_H-768\_A-12. After encoding with BERT, the input text is transformed into indices (index of the word) and segments (which sentence the word belongs to). These two parts are then sent to BERT model and the [CLS] vector is fetched with the Lambda function. This process leads to text embeddings of size 768.

\textbf{Image Encoding}: Images are converted into arrays before encoding. Image features from data are extracted using DenseNet, which builds feature-maps for memes. The image embedding for each meme is the result of the concatenation of feature vectors extracted from the image. The image encoding is then processed into 2 dimensions with Global Max Pooling. The resulting image embeddings are of size 1024. In this procedure, IOError occurs when the system reads images that are corrupted; this error occurred with a single image in the dataset. This was solved by setting LOAD\_TRUNCATED\_IMAGES to true.

The next step is to combine these two embeddings with the concatenate function provided by Keras. After the combination layer, there is a dense layer followed by a dropout layer. There is no need to add too many layers on the top of the model because BERT is so complex that it is competent to extract text features. Besides, the extra layer would have a bad influence on the pre-trained weights of BERT and DenseNet because our added layer will be initialized randomly. If we put several layers on the system, the performance would be reduced, and the model may not convergent. Finally, as the activation functions, we use softmax for task A and task C and sigmoid for task B, which are used for outputting the predicted class. 

\section{Experimental setup}

We tried three different groups of classifiers: (1) unimodal classification using only text (2) unimodal classification using only machine vision, and (3) multimodal classification using text and vision.

\textbf{Data}: We use the publicly released dataset for experiments. As there are eight classification tasks in total, different tasks have different proportions of labels. The details of the splitting train/dev/test are shown in the tables below (Table \ref{taskA}-\ref{taskC}). In total, the training, dev, and test data contain 5,192, 1,800, and 1,878 mentions, respectively.

\begin{table}[htb]
\begin{center}
\begin{tabular}{|l|rrl|}
\hline         &  Train &  Dev   &  Test  \\ \hline
Positive & 3,089 & 1,071 & 1,111     \\
Neutral  & 1,634 & 567   & 594     \\
Negative & 469   & 162   & 173     \\ \hline
Sum      & 5,192 & 1,800 & 1,878 \\
\hline
\end{tabular}
\end{center}
\caption{\label{taskA} Data splitting of task A. }
\end{table}

\begin{table}[htb]
\begin{center}
\begin{tabular}{|c|p{0.8cm}p{0.8cm}p{0.8cm}|p{0.8cm}p{0.8cm}p{0.8cm}|p{0.8cm}p{0.8cm}p{0.8cm}|p{0.8cm}p{0.8cm}p{0.8cm}|}
\hline
    & \multicolumn{3}{c|}{Funny or not} & \multicolumn{3}{c|}{Sarcastic or not} & \multicolumn{3}{c|}{Offensive or not} & \multicolumn{3}{c|}{Motivational or not} \\
    & Train     & Dev       & Test      & Train       & Dev        & Test       & Train       & Dev        & Test       & Train        & Dev         & Test        \\ \hline
Yes & 3,973     & 1,368     & 1,433     & 4,044       & 1,404      & 1,457      & 3,181       & 1,098      & 1,171      & 1,837        & 630         & 1,188       \\
No  & 1,219     & 432       & 445       & 1,148       & 396        & 421        & 2,011       & 702        & 707        & 3,355        & 1,170       & 690         \\ \hline
Sum & 5,192     & 1,800     & 1,878     & 5,192       & 1,800      & 1,878      & 5,192       & 1,800      & 1,878      & 5,192        & 1,800       & 1,878       \\ \hline
\end{tabular}
\end{center}
\caption{Data splitting of task B.}
\label{taskB}
\end{table}

\begin{table}[H]
\begin{center}
\begin{tabular}{|c|ccl|ccl|ccl|}
\hline
\multirow{2}{*}{} & \multicolumn{3}{c|}{Scale of funny} & \multicolumn{3}{c|}{Scale of sarcastic} & \multicolumn{3}{c|}{Scale of offensive} \\
                  & Train      & Dev        & Test      & Train       & Dev         & Test        & Train       & Dev         & Test        \\ \hline
Not               & 1,219      & 432        & 445       & 1,148       & 396         & 421         & 2,011       & 702         & 707         \\
Slightly          & 1,822      & 630        & 654       & 2,607       & 900         & 937         & 1,926       & 666         & 709         \\
Mildly            & 1,662      & 576        & 605       & 1,151       & 396         & 424         & 1,088       & 378         & 387         \\
Very              & 489        & 162        & 174       & 286         & 108         & 96          & 167         & 54          & 75          \\ \hline
Sum               & 5,192      & 1,800      & 1,878     & 5,192       & 1,800       & 1,878       & 5,192       & 1,800       & 1,878       \\ \hline
\end{tabular}
\end{center}
\caption{Data splitting of task C.}
\label{taskC}
\end{table}

\textbf{Parameter settings}: We try different hyper-parameters and chose those which yielded the best result. Specifically, our submitted model is trained in batch mode using the Adam optimizer \cite{Kingma2015} with a learning rate of 1.00E-05. Each batch covers 16 mentions. A dropout rate of 0.2 is applied on the dropout layer and an L2 regularization of 0.02 is used on the dense layer which is adjacent to the last layer. Also, to balance the ratio of different labels and reduce potential overfitting, we add a class weight parameter according to the actual ratio observed in the training data when fitting the model. Additionally, hyper-parameters of comparative experiments (including re-training the BERT-DenseNet model) are shown below in Table \ref{parameter}. Optimizer, batch-size, and class weight are the same as the submitted model.

\begin{table}[H]
\begin{center}
\begin{tabular}{|ll|lll|}
\hline
Type                         & Model         & Learning rate & Dropout & L2 regularization \\ \hline
Text                         & BERT          & 1.00E-06      & 0       & 0.01              \\ \hline
\multirow{2}{*}{Vision}      & DenseNet      & 1.00E-06      & 0       & 0.01              \\
                             & ResNet        & 1.00E-06      & 0       & 0.01              \\ \hline
\multirow{2}{*}{Multi-modal} & BERT-DenseNet & 1.00E-06      & 0.3     & 0.04              \\
                             & BERT-ResNet   & 1.00E-06      & 0.3     & 0.04              \\ \hline
\end{tabular}
\end{center}
\caption{Hyper-parameters of comparative experiments}
\label{parameter}
\end{table}

\textbf{External library}: Keras, the python deep learning library, is used to build the whole model structure. It offers an easy way to load DenseNet\footnote{\url{https://keras.io/applications/#densenet}}, but there is no official method to call BERT. Consequently, keras-bert\footnote{\url{https://github.com/CyberZHG/keras-bert}},an external library and a great implementation of BERT with Keras that could load official pre-trained models, is used in the process of implementing the system.

\textbf{Evaluation}: The organizers proposed to evaluate the performance with the macro-F1 score. For task B and C, the final score is produced by averaging the macro-F1 score for each of the subtasks.

\section{Results}

 Among all participating systems in this SemEval task, our model achieved the 14th score both on the evaluation of task A and task C. Our score on task B is however below the baseline, which we look to improve in future work. Table \ref{result} gives our results, the highest score in this competition and baseline.

\begin{table}[htb]
\centering
\begin{tabular}{|l|lll|}
\hline
Models        & Task A                     & Task B                     & Task C                      \\ \hline
Highest score & \multicolumn{1}{c}{0.3547} & \multicolumn{1}{c}{0.5183} & \multicolumn{1}{c|}{0.3225} \\
BERT-DenseNet & \multicolumn{1}{c}{0.3452} & \multicolumn{1}{c}{0.4421} & \multicolumn{1}{c|}{0.3097} \\
Baseline      & 0.2176                     & 0.5002                     & 0.3009                      \\ \hline
\end{tabular}
\caption{Macro-F1 scores obtained for the highest score model, our final submitted model and baseline.}
\label{result}
\end{table}


Sub-tasks of task B are all binary classification tasks and we expected it to be easier to achieve performance improvements in this case. Surprisingly, however, we get the worst result in task B. Actually, the final submitted result of task B overfits and all of the results are almost positive. But it achieves the best scores among all of our submissions. We believe that changing the parameter to reduce the overfitting effect will yield better results.

Experiments with other models (including re-training the BERT-DenseNet model) are shown below in Table \ref{re-evaluation}. The re-trained BERT-DenseNet model, while ending with lower scores in task A, achieves higher performance than the final submitted model (in Table \ref{result}) in task B and task C. 

\begin{table}[H]
\centering
\begin{tabular}{|ll|lll|}
\hline
Type                         & Models            & Task A          & Task B                              & Task C                      \\ \hline
                             & Highest score     & 0.3547          & \multicolumn{1}{c}{0.5183}          & \multicolumn{1}{c|}{0.3225} \\ \hline
Text                         & BERT              & 0.1574          & 0.4798                              & 0.2749                      \\ \hline
\multirow{2}{*}{Vision}      & \textbf{DenseNet} & \textbf{0.3344} & \multicolumn{1}{c}{\textbf{0.5120}} & \textbf{0.3209}             \\
                             & ResNet            & 0.3186          & 0.4965                              & 0.3129                      \\ \hline
\multirow{2}{*}{Multi-modal} & BERT-DenseNet     & 0.3137          & 0.4999                              & 0.3127                      \\
                             & BERT-ResNet       & 0.3305          & 0.4946                              & 0.3149                      \\ \hline
                             & Baseline          & 0.2176          & 0.5002                              & 0.3009                      \\ \hline
\end{tabular}
\caption{Macro-F1 scores obtained for other models.}
\label{re-evaluation}
\end{table}

We can observe that: (i) DenseNet outperforms ResNet; (ii) introducing BERT would reduce the performance of DenseNet and has little impact on ResNet; (iii) BERT, while yielding SOTA results in a lot of NLP tasks, is not suitable for processing text of memes.

The reasons why DenseNet outperforms ResNet in this SemEval task are two-fold:
\begin{enumerate}
    \item \textbf{Characteristics of DenseNet}: The features extracted by DenseNet are concatenated while ResNet sums the features. Moreover, DenseNets allow layers access to feature-maps from all of its preceding layers. These two characteristics encourage feature reuse throughout the network and lead to more compact models;
    \item \textbf{Dataset}: In total, 6,992 mentions are provided by the organizers. The limited availability of data makes it more difficult to extract sufficient features. Meanwhile, our model is deep. As a result, overfitting may occur easily. Evidence from \cite{huang2017densely} shows that DenseNet is better for small datasets, which is why we opted for this option in our approach.
\end{enumerate}

The results show that DenseNet achieves the best scores in task A among all of our submissions. Therefore, we cannot conclude that uni-modal approaches are better for characterizing internet memes than multi-modal approaches while introducing BERT would reduce the performance of DenseNet. Intuitively, more features encourage models to learn more information from data. We may put more effort into how to connect multi-modal features in that our proposed model concatenates text-based features with image-based features without exploring other options.

\section {Conclusion}

In this paper, we describe our entry for SemEval 2020 Task 8: Memotion Analysis. We present a method for using deep learning to classify memes by emotion. We find that image classification models help classifying memes, with DenseNet outperforming ResNet. Adding text-based features is not always helpful for Memotion Analysis. Our proposed model outperforms baselines except for task B. We are looking to improve performance scores for task B by further tweaking our method and by testing additional parameters.

We should stress that the classifier we trained is still in an early stage of development. We aim to improve it by: (i) enriching the meme datasets from social media and trying to keep the balance among classes; (ii) considering additional features (e.g. adult slang); (iii) trying other methods for combining features. Our present work is limited to one type of meme made of image and text, and we aim to take other types of memes (e.g. video and audio memes) into consideration in the future.

\section {Acknowledgments}

This work was supported by International Cooperation Office of Nanjing University of Aeronautics and Astronautics and College of Economics and Management of Nanjing University of Aeronautics and Astronautics. Xiaoyu Guo conducted part of this work while visiting Queen Mary University of London.

\bibliographystyle{coling}
\bibliography{memes}

\end{document}